\documentclass{article} 
\usepackage{iclr2026_conference,times}

\usepackage{amsmath,amsfonts,bm}

\def\eqref#1{equation~\ref{#1}}

\def\1{\bm{1}}

\DeclareMathAlphabet{\mathsfit}{\encodingdefault}{\sfdefault}{m}{sl}
\SetMathAlphabet{\mathsfit}{bold}{\encodingdefault}{\sfdefault}{bx}{n}

\usepackage{hyperref}
\usepackage[capitalize, noabbrev]{cleveref} 
\usepackage{url}
\usepackage{graphicx} 
\usepackage{booktabs} 
\usepackage{multirow}
\definecolor{mygreen}{rgb}{0.0, 0.5, 0.0}
\usepackage{xcolor}

\title{Seeing Once is Enough? Online Geometry-Aware Token Pruning for 3D Question Answering}

\author{{Ruei-Chi Lai}{$^{1}$}\quad{Bolivar Solarte}{$^{2}$}\quad{Chin-Hsuan Wu}{$^{3}$}\quad{Yi-Hsuan Tsai}{$^{4}$}\quad{Min Sun}{$^{1}$}
\\
{$^{1}$National Tsing Hua University}\quad%
{$^{2}$Industrial Technology Research Institute ITRI}%
\\
{$^{3}$University of Toronto}\quad%
{$^{4}$Atmanity Inc.}%
}

\iclrfinalcopy 
\begin{document}

\maketitle
\vspace{-4mm}
\begin{abstract}
Recent Multi-modal Large Language Models (MLLMs) have demonstrated remarkable performance on 2D question answering tasks.
However, extending these models to the 3D question answering remains challenging, as they typically require multiple views of the scene, which incurs substantial computational cost at inference.
To mitigate this issue, existing solutions rely on strategic frame selection or token-merging algorithms that require preprocessing in advance all frames of the scene, i.e., an offline fashion. In contrast, we propose the first online token-pruning method that can be integrated seamlessly with current MLLM models for 3D question answering tasks, without additional training and with lower memory usage.
Our key insight is to project each input frame into a shared voxel space using depth information and camera pose, identifying
spatially-overlapped regions across frames and selectively pruning redundant image tokens before they enter the language model.
Our method enables efficient online processing while reducing up to 50\% of token usage. We apply this approach to Qwen2.5‑VL‑7B and Qwen3‑VL‑8B, demonstrating improved performance on the ScanQA, SQA3D, and OpenEQA‑HM3D benchmarks.

\begin{figure}[h]
    \centering
    \includegraphics[width=0.8\linewidth]{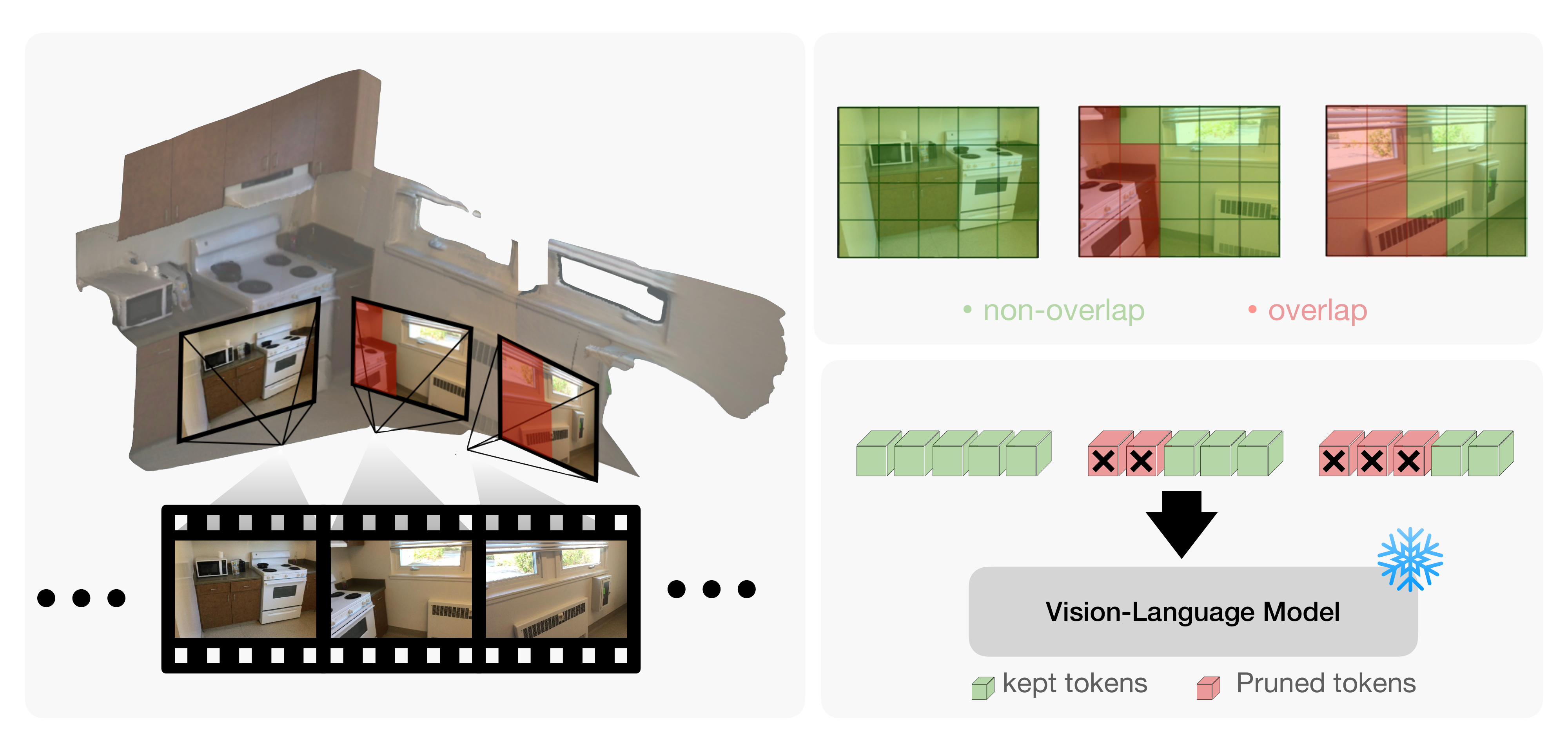}
    \begin{picture}(0, 0)
    
    \put(-311,135){(a)}
    \put(-151,135){(b)}
    \put(-151,69){(c)}
    
    \end{picture}

    \caption{\textbf{Online token pruning.} (a) Sequentially received posed RGB-D frames are voxelized into a shared 3D space. (b) Overlapped regions across frames are identified through voxel comparison. (c) Redundant visual tokens corresponding to these regions are pruned prior to being fed into the VLM, {enabling substantial reduction of token usage without compromising performance.}}
    \vspace{-0.5cm}
    \label{fig:teaser}
\end{figure}
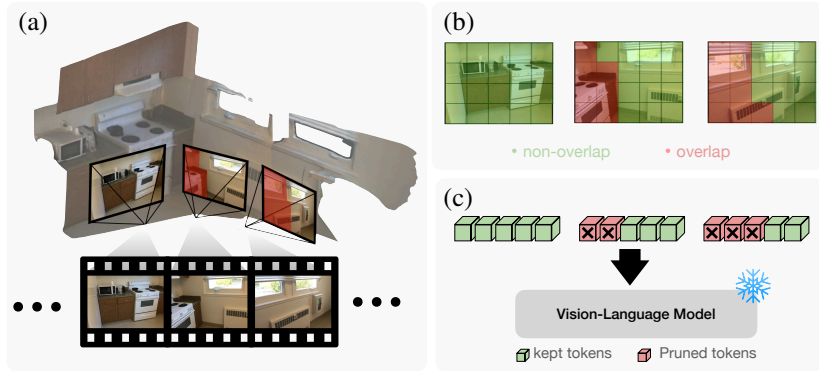

\end{abstract}
\section{Introduction}

Recent advances in Multi-modal Large Language Models (MLLMs) have demonstrated remarkable capabilities in 2D visual question answering~\citep{Qwen2.5-VL, qwen3technicalreport,wang2025internvl3, li2024llava, achiam2023gpt, comanici2025gemini}. However, extending these models to 3D question answering remains challenging. Unlike 2D tasks that rely on a single image, 3D understanding requires processing multiple frames to obtain sufficient spatial information, which greatly increases computational costs and token usage.

Existing efforts have explored two main directions to address 3D question answering. 
First, several works inject 3D priors into MLLMs through specialized 3D modules to enhance spatial reasoning capabilities~\citep{zheng2025video3dllm,xu2024pointllm, huang20253dr1enhancingreasoning3d, Chen_2024_CVPR, tang2024minigpt, Zhu_2025_ICCV, hong20233d, zhang2024chatscene}.
However, these approaches face significant limitations, requiring specialized 3D modules, extensive 3D-language paired datasets, and costly retraining pipelines, which restrict their practical application.
Another line of research aims to reduce visual token usage through strategic frame selection or token compression~\citep{Huang2025CVPR-DTC, zheng2025video3dllm, wu2025spatial, xu2024vlmgrounder}. Although these approaches effectively reduce computational costs and maintain competitive performance, they require access to all images in the scene before inference, making them unsuitable for online tasks such as embodied AI, robotic navigation, or real-time scene understanding, which demand sequential, frame-by-frame processing. This raises the question: \textit{How can we lower computational cost while preserving online processing?}

To this end, we propose the first online, training-free, geometry-aware token pruning method (\cref{fig:teaser}). Our approach leverages depth information and camera poses to project each input frame into a shared 3D voxel space, enabling the tracking of overlapping regions to identify and prune redundant visual tokens.
Remarkably, we find that this geometry-aware pruning not only reduces computational costs but also improves performance on 3D question answering benchmarks, suggesting that removing redundant tokens helps the model to focus on more informative visual cues.

We validate our method on the ScanQA \cite{azuma_2022_CVPR} SQA3D \cite{ma2022sqa3d} and 
OpenEQA-HM3D \cite{majumdar2023openeqa} benchmarks for 3D question answering tasks. We apply our online pruning method on two latest frontier  Vision-Language Models (VLMs), i.e., Qwen2.5-VL-7B and Qwen3-VL-8B without fine-tuning their parameters.
Across all experimental settings, our online pruning strategy reduces token usage by as much as \textbf{50\%} and achieving an improvement of \textbf{+5.1} in the \textbf{LLM-Match} score.
Our contributions are summarized as follows:

\begin{itemize}
    \item We propose the first training-free, geometry-aware 
    token pruning method for 3D question answering that can be seamlessly integrated into existing 2D VLMs and run in an online manner.

    \item Extensive experiments demonstrate that our pruning method can substantially reduce token usage while consistently improving performance across all settings.
    
    \item We further evaluate our approach across multiple models and benchmarks (ScanQA, SQA3D, OpenEQA-HM3D), demonstrating both its strong generalization capabilities and its effectiveness for 3D question answering tasks.
    
\end{itemize}

\section{Related Work}
\label{sec:related_work}

\subsection{3D MLLMs}
{
With the rapid development of Multi-modal Large Language Models (MLLMs), numerous works~\citep{Chen_2024_CVPR, chen2024grounded3dllm, wang2023chat, huang2024embodied, zhang2024chatscene, GPT4Scene, Zhu_2025_ICCV, xu2024pointllm, zheng2025video3dllm, huang20253dr1enhancingreasoning3d} recently boost the MLLMs with 3D scene understanding capabilities using 3D information such as 3D point clouds and 3D bounding boxes. LL3DA~\citep{Chen_2024_CVPR} and Grounded 3D-LLM~\citep{chen2024grounded3dllm} leverage 3D detectors or segmentation modules to extract object-centric features for language reasoning. Chat3D~\citep{wang2023chat}, LEO~\citep{huang2024embodied}, and Chat-Scene~\citep{zhang2024chatscene} adopt a similar approach by encoding segmented 3D objects and fusing their features into large language models. GPT4Scene~\citep{GPT4Scene} constructs a bird's-eye-view (BEV) image by reconstructing the 3D scene for question answering. LLaVA-3D~\citep{Zhu_2025_ICCV} integrates 3D geometric position information into image patch embeddings with instruction tuning to align 2D and 3D modalities, while Video-3D LLM~\citep{zheng2025video3dllm} employs a 3D positional encoding module with post-instruction tuning for 3D scene understanding. However, these methods depend on 3D training data, which is scarcer than large-scale 2D image datasets. The scarcity of high-quality 3D resources poses a bottleneck for scaling 3D MLLMs, limiting their coverage of real-world scene variations and capacity to learn generalizable spatial reasoning, highlighting the need for more data-efficient approaches.

}

\subsection{3D Question Answering with VLMs}
{
Recent advancements in Vision-Language Models (VLMs) have extended multimodal reasoning from 2D images to multi-frame and 3D scene understanding~~\citep{Liu_2025_CVPR, li2024llava, Qwen2.5-VL, qwen3technicalreport, comanici2025gemini, achiam2023gpt, lin2024video}. Benchmarks such as ScanQA~\citep{azuma_2022_CVPR}, SQA3D~\citep{ma2022sqa3d}, and OpenEQA~\citep{majumdar2023openeqa} evaluate 3D question answering, while ScanRefer~\citep{chen2020scanrefer} and Multi3DRefer~\citep{Zhang_2023_ICCV} focus on 3D grounding. Unlike grounding, 3D question answering requires spatial and semantic reasoning across multiple views to answer scene-level questions.

Several VLM families, including LLaVA~\citep{li2024llava}, Video-LLaVA~\citep{lin2024video}, and Qwen-VL~\citep{Qwen2.5-VL, qwen3technicalreport}, have demonstrated strong cross-modal understanding through large-scale instruction tuning. However, most existing VLMs remain inherently 2D-based, relying solely on image sequences without explicitly incorporating 3D geometric information, limiting their spatial reasoning in complex 3D environments. In this work, we bridge this gap by exploring a balance between utilizing 3D spatial information and leveraging the strong generalization capability of 2D VLMs for effective 3D scene question answering.

}

\subsection{Frame Sampling Strategy}
{
To enable MLLMs to understand 3D scenes through powerful visual reasoning, visual-based 3D scene understanding typically relies on video frames as input. However, due to limited GPU memory, models can only process a subset of frames, even though a single 3D scene dataset may contain thousands of frames. A widely adopted approach is uniform frame sampling~\citep{Qwen2.5-VL, yang2025thinking, zheng2025video3dllm, Huang2025CVPR-DTC}, which evenly selects frames from the entire sequence. While simple and efficient, this strategy often reaches its performance ceiling.

To overcome this limitation, several works~\citep{zheng2025video3dllm, wu2025spatial} have leveraged 3D geometric information and adopted greedy algorithms to select frames that maximize spatial coverage while minimizing the total number of frames. Some prior works~\citep{Hu_2025_CVPR} propose trainable frame selectors that predict frame importance scores for query-relevant frame identification. Similarly, VLM-Grounder~\citep{xu2024vlmgrounder} introduces a query-aware frame selection framework, while Dynamic Token Compression (DTC)~\citep{Huang2025CVPR-DTC} presents an offline token reduction method combining 3D voxelization with visual similarity matching to reduce visual tokens while maintaining performance.

However, most existing strategies rely on offline processing or additional selector modules, increasing computational overhead and making them unsuitable for online or real-time scenarios. We propose an online, single-pass frame sampling strategy that leverages 3D geometric information to reduce redundant tokens while improving model performance, effectively balancing efficiency and 3D scene understanding in 3D question answering tasks.
}

\section{Proposed Method}
\label{sec:method}

\begin{figure}[t]
    \centering
    \includegraphics[width=1\linewidth]{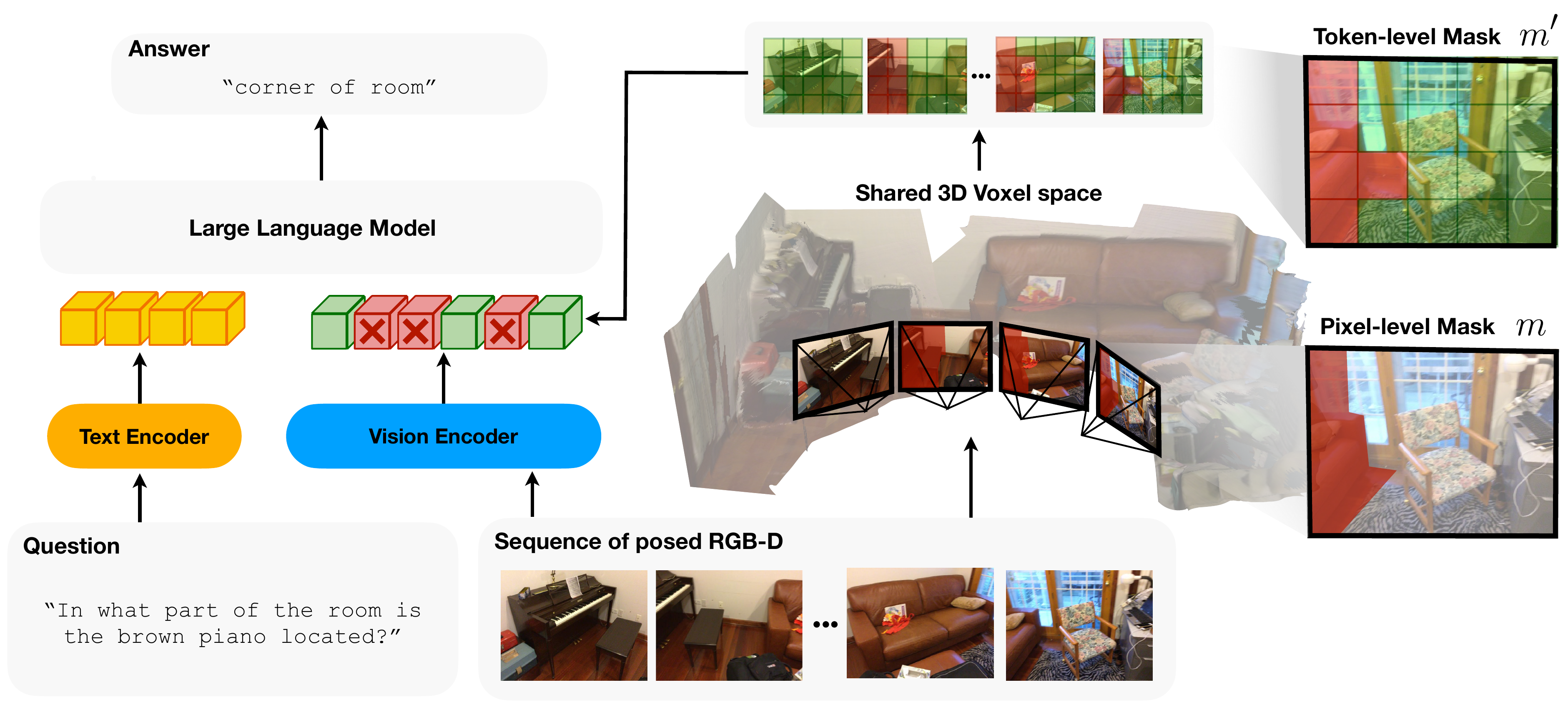}
    \caption{\textbf{Pipeline of our method.} Our method processes frames sequentially via online 3D voxelization, generating a pixel-level mask $m$ from overlapped regions and computing a token-level mask $m'$ to effectively prune redundant tokens before they enter the LLM. 
    }
    \label{fig:our_method}
    \vspace{-0.5cm}
\end{figure}

In this section, we introduce our online geometry-aware token pruning method for the 3D question answering task, which aims to maximize 3D scene information while remaining computationally efficient. For reference purposes, the overall pipeline is illustrated in~\cref{fig:our_method}.
In~\cref{sec:frame_sampling}, we discuss frame sampling strategies under both online and offline settings.
In~\cref{sec:3d_projection}, we present our method to identify redundant information in overlapping regions under a sequential frame input setting.
Finally, in~\cref{sec:token_pruning}, we describe how redundant information is dropped during MLLMs' inference without requiring any further post-training.

\subsection{Frame Sampling}
\label{sec:frame_sampling}

Processing scene videos in VLMs requires selecting a subset of frames due to limited GPU memory and computational resources. An effective frame sampling strategy should preserve the essential spatial coverage of the 3D scene while minimizing redundancy among frames. Existing approaches typically perform offline frame selection, where the entire video or dataset must be available beforehand to optimize the coverage. However, this assumption is impractical for streaming or real-time scenarios, where frames are observed sequentially and future content is unknown. To address this limitation, we first consider uniform sampling as a online strategy that does not require pre-processing the entire set of video frames. However, uniform sampling fails to account for spatial redundancy between frames and often leads to suboptimal scene coverage, motivating the need for a more adaptive online selection mechanism.

\paragraph{Uniform sampling.}

Uniform sampling intuitively selects frames from the video frame set ${F} = \{f_1, f_2, ..., f_n\}$ at a fixed interval of every ${n}/t$ where ${t}$ denotes the total number of frames to be sampled. However, uniform sampling introduces a critical trade-off between temporal coverage and computational efficiency. When the sampling interval is large, the model may skip important frames containing fine-grained spatial or semantic details of the 3D scene. In contrast, reducing the interval to capture more information leads to a larger number of selected frames, quickly exceeding GPU memory limits. As a result, uniform sampling often fails to achieve an optimal balance between efficiency and scene coverage.

\paragraph{Maximum coverage sampling.}

We follow the offline maximum coverage sampling strategy introduced in prior work~\citep{zheng2025video3dllm}, which aims to select the fewest possible frames while maximizing the coverage of the 3D scene. This problem is formulated as a maximum coverage problem and is known to be NP-hard. Therefore, a greedy algorithm is adopted to obtain an approximate solution. Specifically, given a video with frame set ${F} = \{f_1, f_2, ..., f_n\}$, each frame is projected into 3D voxels, and the goal is to find a subset of frames ${S} \subseteq {F}$ that maximizes the overall voxel coverage.
\subsection{3D Projection}
\label{sec:3d_projection}
For each incoming posed RGB-D image, we first project the image into the 3D world coordinate system, obtaining ${C} \in \mathbb{R}^{H\times W\times 3}$ for every pixel using the corresponding depth map and camera pose. Each pixel $(u,v)$ is thereby associated with a 3D point ${C}(u,v) = (x,y,z)$ in the global coordinate space. The resulting 3D points are then voxelized to produce discrete voxel indices $\mathcal{V} \in \mathbb{Z}^{(H\cdot W)\times 3}$, which represent quantized spatial locations in the scene.

To determine whether a spatial region has been previously observed, we maintain a global voxel set $\mathbf{S}$ that records all voxels visited by past frames. 
For each new frame, voxels corresponding to $\mathcal{V}$ are compared against $\mathbf{S}$; if a voxel already exists in $\mathbf{S}$, it is considered overlapped. These overlapped voxels are subsequently back-projected onto the image plane to produce a 
pixel-level binary mask ${m} \in \{0,1\}^{H\times W}$, where ${m}(u,v)=1$ indicates that the
corresponding pixel is projected from a region that has already been observed. This mask effectively captures spatial redundancy across frames at the pixel level.

To bridge the pixel-level redundancy and the visual tokens processed by the vision encoder, the mask ${m}$ is spatially aggregated over non-overlapping patches of size $P\times P$ to form a token-level mask ${m'} \in \{0,1\}^{H'\times W'}$, where $H'=\left\lfloor \tfrac{H}{P} \right\rfloor$ and $W'=\left\lfloor \tfrac{W}{P} \right\rfloor$. Each patch corresponds to one visual token in the encoder, and we compute an overlap ratio for each token as:
\begin{equation}
r_{i,j} = \frac{1}{|\mathcal{B}{i,j}|}
\sum_{(u,v)\in\mathcal{B}_{i,j}} {m}(u,v),
\label{eq:overlap_ratio}
\end{equation}
where $\mathcal{B}{i,j}$ denotes the set of pixels within the $(i,j)$\textsuperscript{th} patch.
The overlap ratio $r_{i,j}$ measures the proportion of pixels within a token region that have been previously observed, serving as a quantitative indicator of visual redundancy.

We then define the token-level binary mask as:
\begin{equation}
m'_{i,j} =
\begin{cases}
1, & \text{if } r_{i,j} \ge \tau_o, \\
0, & \text{otherwise},
\end{cases}
\label{eq:token_mask}
\end{equation}
where $\tau_o$ is the pruning threshold that specifies the minimum proportion of overlap required for a token to be considered redundant. In practice, this threshold provides a controllable trade-off between pruning aggressiveness and information retention. A higher $\tau_o$ prunes only highly redundant tokens, while a lower value removes tokens more aggressively.

The resulting mask ${m'}$ is later used to guide the pruning process of visual tokens before feeding them into the large language model. This mechanism ensures that only novel or spatially informative regions are retained, thereby reducing token redundancy and improving computational efficiency. Further analysis of the impact of different pruning thresholds is presented in the experimental section.
\subsection{Token Pruning}
\label{sec:token_pruning}

The RGB image is first fed into the vision encoder to generate a grid of visual tokens
$e \in \mathbb{R}^{H' \times W' \times d}$, where $d$ denotes the token embedding dimension.
Each token $e_{i,j}$ corresponds to a localized spatial region in the input frame and encodes its visual information. 

To identify redundant information across frames, we leverage the 3D projection-based overlap analysis introduced in~\cref{sec:3d_projection}. The mask $m'$ reflects which spatial regions have already been observed or overlapped by previously processed frames. Specifically, $m'_{i,j} = 1$ indicates that the corresponding token $e_{i,j}$ belongs to an overlapped region that conveys redundant visual content and should therefore be pruned. Conversely, $m'_{i,j} = 0$ denotes that the token originates from a novel or unobserved region and should be retained for feeding into the LLM. The pruned token set can thus be formulated as:
\begin{equation}
e' = \{\, e_{i,j} \mid m'_{i,j} = 0 \,\},
\end{equation}
where $e'$ represents the subset of tokens that contribute unique spatial information. When feeding $e'$ into the LLM, redundant tokens are excluded from both the prefill and forward phases, reducing computational overhead and mitigating attention dilution.  
This geometry-aware pruning removes visual tokens from already-explored regions, effectively reducing redundancy while preserving essential context. As a result, the VLM focuses more on novel spatial cues, improving both efficiency and 3D question answering performance.


\section{Experiments}

\subsection{Experimental Setup}
\label{sec:experimental_setup}
\paragraph{Implementation details.}
We use Qwen2.5-VL-7B and Qwen3-VL-8B as base models. Two sampling strategies are adopted: uniform sampling and maximum coverage (MC) sampling. For uniform sampling, we select 20 frames on the ScanQA~\citep{azuma_2022_CVPR}, SQA3D~\citep{ma2022sqa3d}, and OpenEQA-HM3D~\citep{majumdar2023openeqa} benchmarks. As discussed in ~\cref{sec:frame_sampling}, MC sampling adaptively selects frames, averaging around 20 per scene. To ensure a fair comparison in token usage, we uniformly sample 30 frames for all benchmarks before applying our method, while using the same selected frames for offline MC. The pruning threshold is set to 100\%, images are resized to 640×480, and voxel size is 0.1 m. We will release the code to the public.

\paragraph{Datasets \& evaluation metrics.}
We conduct our experiments on three 3D question answering datasets: ScanQA-val, SQA3D-test, and OpenEQA-HM3D to evaluate the effectiveness of our geometry-aware token pruning method. ScanQA and SQA3D are built upon the ScanNet dataset~\citep{dai2017scannet}. Both emphasize evaluating MLLMs' spatial understanding capabilities in 3D scenes. ScanQA-val includes 71 distinct scenes and 4,675 questions, while SQA3D-test includes 67 distinct scenes and 3,519 questions. Different from the ScanNet dataset, OpenEQA-HM3D is built on the HM3D dataset, which comprises real-world questions and focuses on embodied question answering. OpenEQA-HM3D includes 63 distinct scenes and 557 questions. 

In this study, we follow prior works~\citep{azuma_2022_CVPR, hong20233d, zheng2025video3dllm, wu2025spatial} and report Exact Match and CIDEr scores for the ScanQA and SQA3D benchmarks. We also follow OpenEQA and report the GPT-4.1-mini LLM-Match score in our experiments. We further include the total token consumption required to evaluate the full benchmark, reported as ``Tokens'' in the table.

\subsection{Experimental Results}
\label{sec:results}
\begin{table*}[ht]
\centering
\fontsize{6.5}{10}\selectfont
\centering
\caption{\textbf{Overall performance comparisons for the online strategy.} ``Fine-tuned'' indicates that the model is trained on the ScanQA and SQA3D datasets.}
\vspace{1mm}
\begin{tabular}{l c c ccc | cc | cc}
    \toprule
    \multirow{2}{*}{Model} 
    & \multirow{1}{*}{Online}
    & \multirow{1}{*}{Sampling}
    & \multicolumn{3}{c}{ScanQA}
    & \multicolumn{2}{c}{SQA3D}
    & \multicolumn{2}{c}{OpenEQA-HM3D}
    \\
    \cmidrule(lr){4-6} \cmidrule(lr){7-8} \cmidrule(lr){9-10}
    & Sampling & Strategy & EM$~\uparrow$& CIDEr$~\uparrow$ & Tokens$~\downarrow$ & EM$~\uparrow$ & Tokens$~\downarrow$ & LLM Match$~\uparrow$ & Tokens$~\downarrow$\\
    \midrule
    \textbf{\textit{Fine-tuned 3D LLMs}} \\
    \multirow{1}{*}{Video-3D LLM}
    & \checkmark & Uniform
    & 29.6 & 99.6 & - & 58.3 & - & - & -   \\
    \midrule
    \textbf{\textit{Zero-shot 2D VLMs}} \\
    \multirow{2}{*}{Qwen2.5-VL-7B}
    & \checkmark & Uniform
    & 24.1 & 65.6 & 37.1M& 46.5 & 28.0M& 53.1 & \textbf{4.4M}\\
    & \checkmark & Uniform + Ours 
    & \textbf{25.1} & \textbf{69.3} & \textbf{32.6M}& \textbf{47.3} & \textbf{24.1M}& \textbf{58.2} & 5.2M   \\
    
    \cmidrule(lr){1-10}
    
    \multirow{2}{*}{Qwen3-VL-8B} 
    & \checkmark & Uniform
    & 27.2 & 78.7 & 28.6M& 50.2 & 21.5M& 67.1 & \textbf{3.4M}\\
    & \checkmark & Uniform + Ours 
    & \bf{27.9} & \bf{79.9} & \textbf{26.2M}& \bf{50.7} & \textbf{19.4M}& \bf{67.8} & 4.1M\\

    \bottomrule
\end{tabular}
\label{tab_main_result_online}
\vspace{3mm}
\end{table*}
\begin{table*}[ht]
\centering
\fontsize{7}{10}\selectfont
\centering
\caption{\textbf{Overall performance comparisons for offline strategy.} We report DTC EM score when retaining 54\% of the tokens with 12 uniformly sampled frames. ``Fine-tuned'' indicates that the model is trained on the ScanQA and SQA3D datasets. }
\vspace{1mm}

\begin{tabular}{l c c ccc | cc}
    \toprule
    \multirow{2}{*}{Model} 
    & \multirow{1}{*}{Online}
    & \multirow{1}{*}{Sampling}
    & \multicolumn{3}{c}{ScanQA}
    & \multicolumn{2}{c}{SQA3D}
    \\
    \cmidrule(lr){4-6} \cmidrule(lr){7-8}
    & Sampling 
    & Strategy
    & EM$~\uparrow$ & CIDEr$~\uparrow$ & Tokens$~\downarrow$ & EM$~\uparrow$ & Tokens$~\downarrow$\\
    \midrule
    \textbf{\textit{Fine-tuned 3D LLMs}} \\
    \multirow{1}{*}{Video-3D LLM}
    & & MC
    & 29.8 & 100.3 & - & - & - \\
    \midrule
    \textbf{\textit{Zero-shot 2D VLMs}} \\
    \multirow{1}{*}{LLava-OV-7B}
    & & DTC
    & 27.8 & - & - & -  & - \\
    \cmidrule(lr){1-8}

    \multirow{2}{*}{Qwen2.5-VL-7B}
    & & MC 
    & \textbf{26.2} & 70.9 & 38.6M& \textbf{48.2} & 28.5M \\
    & & MC + Ours 
    & 26.1 & \textbf{71.4} & \textbf{28.3M}& 47.8 &\textbf{20.8M} \\
    
    \cmidrule(lr){1-8}
    
    \multirow{2}{*}{Qwen3-VL-8B} 
    & & MC 
    & 28.0 & 80.1 & 29.8M& 50.5 & 22.0M  \\
    & & MC + Ours
    & \bf{28.2} & \bf{80.5} & \textbf{22.5M}& \bf{51.1} & \textbf{16.6M}\\
    
    \bottomrule
    
\end{tabular}

\label{tab_main_result_offline}
\vspace{3mm}
\end{table*}
\begin{table}[ht]
\centering
\fontsize{7}{10}\selectfont
\caption{\textbf{Categorical results on SQA3D.} Comparisons between online and offline sampling strategies, with evaluation results reported across six categories of SQA3D. Our method can be integrated with both strategies and enhances them.}
\centering
\begin{tabular}{l c c ccccccc c}
    \toprule
    \multirow{2}{*}{Model} 
    & \multirow{1}{*}{Online}
    & \multirow{1}{*}{Sampling}
    & \multicolumn{8}{c}{SQA3D}
    \\
    \cmidrule(lr){4-11}
    & Sampling & Strategy & what&which&can&is&how&others&Average$~\uparrow$& Tokens$~\downarrow$ \\
    \midrule

    \multirow{4}{*}{Qwen2.5-VL-7B} 
    & \checkmark & Uniform
    & 41.2 & 45.9 & \textbf{56.5} & \textbf{58.3} & 37.4 & 45.8 & 46.5 & 28.0M \\
    & \checkmark & Uniform + Ours
    & \textbf{41.7} & \textbf{49.3} & 53.6 & 57.8 & \textbf{41.1} & \textbf{46.8} & \textbf{47.3} & \textbf{24.1M} \\
    \cmidrule(lr){2-11}
    &  & MC
    & \textbf{42.0} & \textbf{49.6} & \textbf{56.2} & \textbf{58.1} & \textbf{43.0} & \textbf{48.1} & \textbf{48.2} & 28.5M \\
    &  & MC + Ours
    & 41.8 & 49.0 & 55.3 & \textbf{58.1} & 41.7 & 47.9 & 47.8 & \textbf{20.8M} \\
    
    \midrule
    
    \multirow{4}{*}{Qwen3-VL-8B} 
    & \checkmark & Uniform
    & 45.6 & 45.6 & 54.1 & 60.6 & 46.2 & 51.1 & 50.2 & 21.6M\\
    & \checkmark & Uniform + Ours
    & \textbf{46.4} & \textbf{48.2} & \textbf{55.6} & \textbf{60.9} & \textbf{44.3} & \textbf{51.4} & \textbf{50.7} & \textbf{19.4M} \\
    \cmidrule(lr){2-11}
    &  & MC
    & 46.7 & 41.9 & \textbf{57.7} & 61.2 & 46.0 & 50.5 & 50.5 & 22.0M \\
    &  & MC + Ours 
    & \textbf{47.2} & \textbf{42.5} & 56.5 & \textbf{61.8} & \textbf{47.1} & \textbf{51.9} & \textbf{51.1} & \textbf{16.6M}\\
    
    \bottomrule
\end{tabular}

\label{tab_sqa3d_category}
\vspace{3mm}
\end{table}

We present the overall comparison between leveraging uniform sampling and maximum coverage sampling strategies, as well as applying our method, in both \cref{tab_main_result_offline} and \cref{tab_main_result_online}. Different from fine-tuned models that post-trained on downstream tasks, we use zero-shot VLMs such as Qwen2.5-VL-7B and Qwen3-VL-8B as our base models.

\vspace{-3mm}
\paragraph{Online scenario.}
In \cref{tab_main_result_online}, we compare our method with the online uniform sampling strategy, which uses 20 frames for ScanQA, SQA3D, and OpenEQA-HM3D.
To ensure a fair comparison in token usage, we apply our method on 30 uniformly sampled frames for all benchmarks, while pruning overlapped visual tokens to match the baseline’s token budget. As shown in the table, when applying our method with more frames, the redundant overlapped tokens identified by our pruning process are effectively removed. This allows the model to focus more on the informative regions, not only reducing token usage but also improving overall performance. The effectiveness of our method is consistently demonstrated on both QwenVL-series models, where performance improvements are observed across three benchmarks. Specifically, on Qwen2.5-VL-7B, our approach improves the Exact Match score from 24.1 to 25.1 on ScanQA and from 46.5 to 47.3 on SQA3D, while using fewer tokens. The results indicate that when the online setting is required, our method provides a more effective sampling strategy without introducing any preprocessing or selector modules that would increase computational overhead. To further analyze scalability, we present a detailed comparison of different numbers of uniformly sampled frames and the effect of applying our method in \cref{sec:ablation_study}, as illustrated in \cref{fig:ablation_2}.

\vspace{-3mm}
\paragraph{Offline scenario.}
In \cref{tab_main_result_offline}, we compare our method with the offline sampling strategies. We include prior work DTC~\citep{Huang2025CVPR-DTC} in our comparison. DTC compresses visual tokens to reduce the input length, we report its EM score when retaining 54\% of the tokens. In our evaluation setup, we adopt the adaptive maximum coverage (MC) sampling strategy, which selects approximately 20 frames per question across all benchmarks. As shown in the table, even under the same set of offline-processed frames—where MC is used to select the most representative and diverse frames—our method is able to significantly reduce token usage while maintaining competitive performance. This makes our approach particularly suitable for scenarios with limited computational resources. Furthermore, applying our method achieves comparable Exact Match scores on both ScanQA and SQA3D, improves the CIDEr score on ScanQA with Qwen2.5-VL-7B. These results demonstrate that our pruning strategy effectively removes redundant visual tokens, optimizing the input representation in a way that serves as an efficient offline enhancement to existing sampling methods.

\vspace{-3mm}
\paragraph{Categorical results.}
In \cref{tab_sqa3d_category}, we present the evaluation results across six categories of the SQA3D benchmark for both QwenVL-series models. For the online sampling strategies, the results demonstrate that our method not only improves the overall performance but also yields consistent gains across almost all categories. In the offline setting, our method maintains comparable performance across all six categories. Notably, for Qwen3-VL-8B, applying our method on top of the MC baseline achieves a 1.4 points improvement in Exact Match (EM) for the “others” category. Across both settings, our method consistently reduces the total number of visual tokens, highlighting its efficiency in balancing performance and token usage.

\subsection{Ablation Study}
\label{sec:ablation_study}
\begin{figure}[h]
  \centering
   \includegraphics[width=0.6\linewidth]{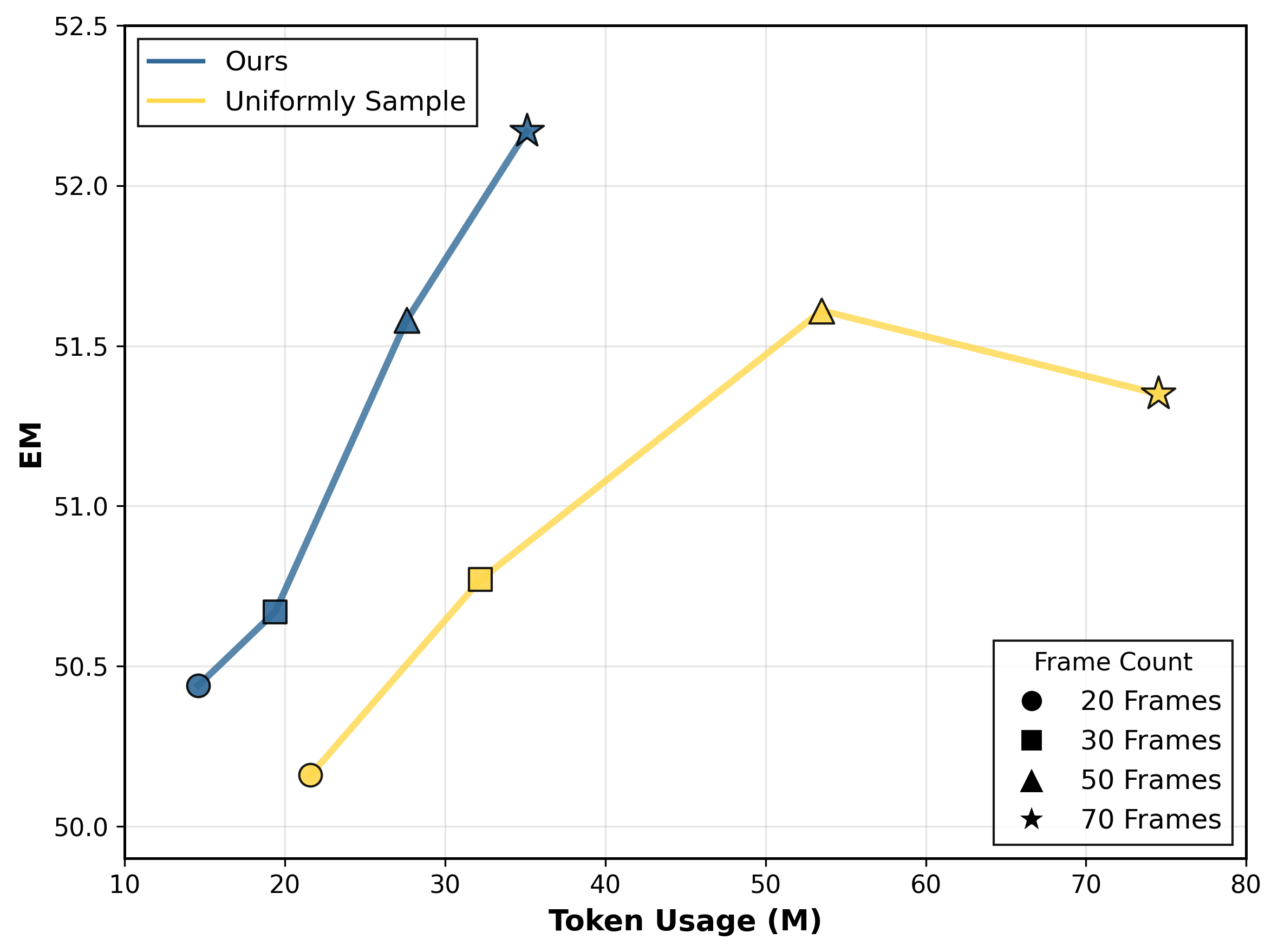}
    \caption{\textbf{Ablation on scaling the number of frames.} Our method maintains stable or improved performance while significantly reducing token usage compared to uniform sampling.}
   \label{fig:ablation_2}
\end{figure}

\paragraph{Scaling up sampled frames.}
In \cref{fig:ablation_2}, We investigate the scalability of our method when increasing the number of sampled frames. From the figure, we observe that under the same 20-frame setting, our approach achieves better performance while using fewer visual tokens compared to the baseline uniform sampling strategy. As the number of frames increases to 30 and 50, our method consistently removes redundant tokens while maintaining or slightly improving Exact Match (EM) scores. When input frames scale up to 70, the baseline performance degrades due to excessive redundancy, while our method continues to yield stable or improved performance with significantly reduced token count. This indicates that our geometry-aware pruning effectively filters out repetitive visual information, allowing the model to focus on novel and spatially informative regions even when frame input becomes dense. This finding is particularly important in large-scale 3D environments, where comprehensive scene coverage often demands more input frames. While increasing frames typically causes quadratic growth in token usage and computational cost, our approach maintains a balance between accuracy and efficiency by pruning tokens while preserving critical scene context.

\begin{table}[ht]
\centering
\fontsize{8}{11}\selectfont
\centering
\caption{\textbf{Ablation study on the pruning threshold~$\tau_o$.} We investigate how different threshold settings affect pruning behavior, with higher $\tau_o$ values leading to less aggressive token removal. The results demonstrate that a threshold of 100\% performs the best.}
\vspace{1mm}
\begin{tabular}{l c c ccccccc c}
    \toprule
    \multirow{2}{*}{Model} 
    & \multirow{2}{*}{Sampling Strategy}
    & \multirow{2}{*}{$\tau_o$}
    & \multicolumn{2}{c}{SQA3D}
    \\
    \cmidrule(lr){4-5}
    & & & EM$~\uparrow$ & Tokens$~\downarrow$ \\
    \midrule
    \multirow{5}{*}{Qwen3-VL-8B} 
    
    & Uniform & - & 51.4 & 74.5M \\
    \cmidrule{2-5}
    & Uniform + Ours& 25\% 
    & 49.6 & \textbf{10.7M} \\
    & Uniform + Ours & 50\% 
    & 50.8 & 13.5M\\
    & Uniform + Ours & 80\% 
    & 50.8 & 19.4M \\
    & Uniform + Ours & 100\% 
    & \bf{52.2} & 35.2M\\ 
    
    \bottomrule
\end{tabular}

\label{tab_ablation_1_threshold}
\vspace{3mm}
\end{table}
\paragraph{Pruning threshold.}

In \cref{eq:overlap_ratio} and \cref{eq:token_mask}, we introduce the pruning threshold $\tau_o$, specifying the minimum proportion of overlapped pixels within a patch required for the corresponding image token to be considered redundant and pruned. This ensures that a token is removed only when a sufficiently large portion of its spatial region has already been observed. We evaluate the effect of different pruning thresholds on the SQA3D dataset, as shown in \cref{tab_ablation_1_threshold}. Using 70 uniformly sampled frames as the baseline, we observe that the pruning threshold $\tau_o$ plays a crucial role in shaping the model's behavior. When a lower threshold is applied, model performance tends to decrease as more regions are considered redundant and aggressive pruning takes effect. In contrast, when the threshold reaches 100\%—meaning all pixels within a patch are identified as overlapped—the model achieves better performance than the baseline, improving from 51.4 to 52.2 while using only 47\% of the tokens. These results suggest that properly controlling the pruning threshold enables the system to effectively filter out redundant tokens, allowing the model to focus on novel and spatially informative visual regions rather than repeated observations.

\subsection{Qualitative Result}
\label{sec:qualitative_result}

\begin{figure}[h]
    \centering
    \includegraphics[width=1\linewidth]{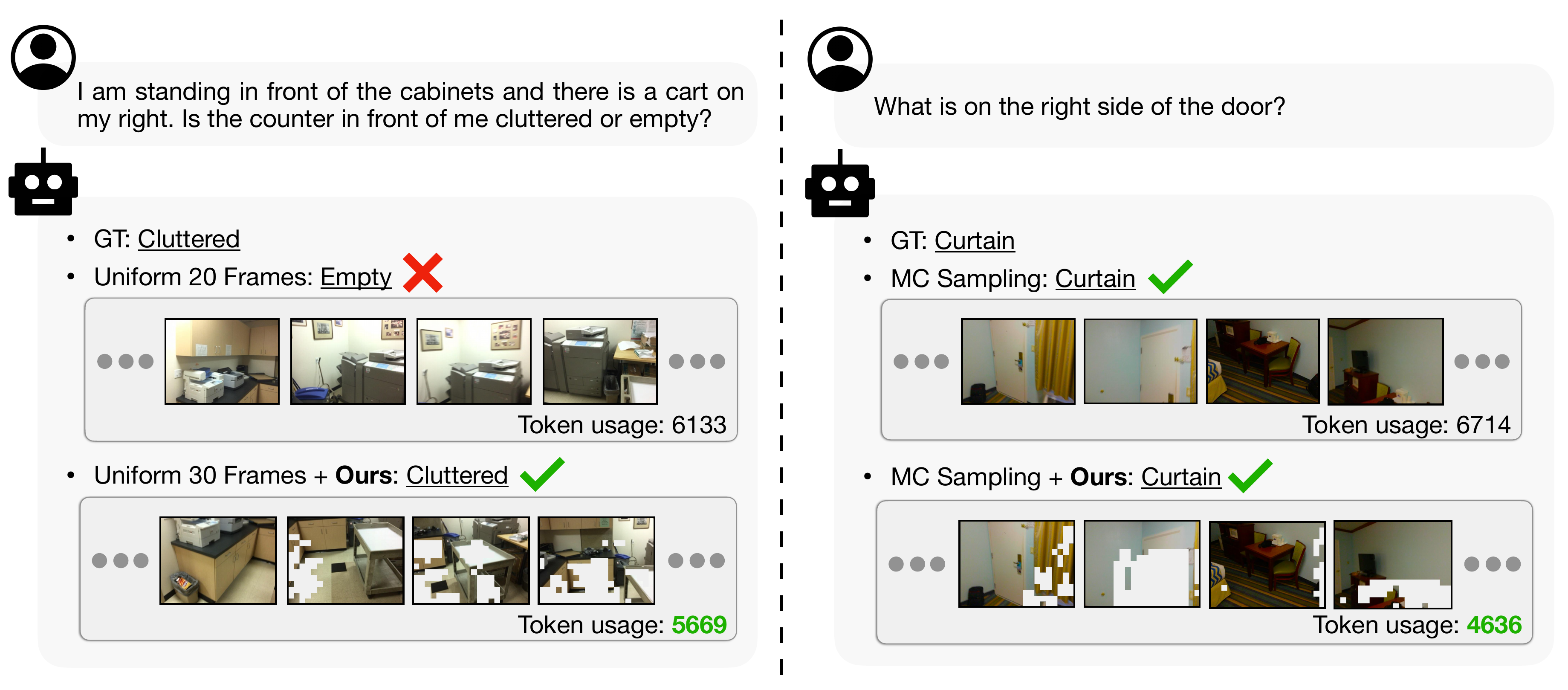}
    \caption{\textbf{Qualitative results.} We adopt our method with two sampling strategies: uniform (left) and MC (right), on SQA3D and ScanQA, respectively. Our method consistently improves token efficiency and performance. }
    
    \label{fig:qualitative_result}
\end{figure}

In ~\cref{fig:qualitative_result}, we present two visualization examples demonstrating our method's effectiveness under online uniform sampling and offline maximum coverage sampling. The left half compares uniform sampling with 20 frames against our method applied to 30 uniformly sampled frames. Despite processing more frames initially, our method significantly reduces the final token count through geometry-aware pruning. More importantly, this token reduction improves answer accuracy, while the baseline produces an incorrect response, our method successfully removes redundant visual information, allowing the model to focus on relevant spatial features and generate the correct answer. The right half shows our method combined with maximum coverage sampling. Both approaches produce correct answers, validating that our pruning preserves essential visual information. However, our method reduces token consumption from 6.7k to 4.6k tokens while maintaining identical performance. This substantial reduction in computational cost without sacrificing accuracy demonstrates that our method successfully identifies and eliminates redundant visual tokens.

\section{Conclusion and Future Work}
\label{sec:conclusionandfuturework}
In this study, we propose an online, single-pass geometry-aware token pruning method that enhances 3D question answering performance while reducing visual token usage. Our approach projects image pixels into a shared voxel space to identify overlapped regions and generates a pruning mask to remove redundant tokens before feeding into the large language model, allowing the model to focus on novel and informative regions in 3D scenes.

Extensive experiments demonstrate that our method outperforms the online uniform sampling baseline on ScanQA, SQA3D, and OpenEQA-HM3D. Compared with the offline maximum coverage strategy, our method reduces token usage from 29.8M to 22.5M on Qwen3-VL-8B for ScanQA while improving both Exact Match and CIDEr scores, providing an efficient and effective solution for scalable 3D scene understanding in VLMs.

While our framework improves performance and reduces token usage, some limitations remain. Pruned tokens from overlapped regions may still contain useful spatial cues, and future work could explore integrating this information or combining geometry-aware pruning with token compression to further enhance efficiency. Additionally, our method is currently evaluated only on predefined datasets, and future work will explore applying our approach within embodied AI systems to handle real-time, real-world scenarios.

\bibliography{iclr2026_conference}

@String(CVPR= {IEEE Conf. Comput. Vis. Pattern Recog.})

@String(ICCV= {Int. Conf. Comput. Vis.})

@String(ECCV= {Eur. Conf. Comput. Vis.})

@String(CVPR  = {CVPR})

@String(ICCV  = {ICCV})

@String(ECCV  = {ECCV})

@inproceedings{zheng2025video3dllm,
  title        = {Video-3D LLM: Learning Position-Aware Video Representation for 3D Scene Understanding},
  author       = {Duo Zheng and Shijia Huang and Liwei Wang},
  booktitle    = {Proceedings of the IEEE/CVF Conference on Computer Vision and Pattern Recognition (CVPR)},
  year         = {2025},
  note         = {arXiv preprint arXiv:2412.00493},
  url          = {https://arxiv.org/abs/2412.00493}
}

@InProceedings{Huang2025CVPR-DTC,
  author    = {Huang, Hsiang-Wei and Chen, Fu-Chen and Chai, Wenhao and Su, Che-Chun and Xia, Lu and Jung, Sanghun and Yang, Cheng-Yen and Hwang, Jenq-Neng and Sun, Min and Kuo, Cheng-Hao},
  title     = {Zero-shot 3D Question Answering via Voxel-based Dynamic Token Compression},
  booktitle = {Proceedings of the IEEE/CVF Conference on Computer Vision and Pattern Recognition (CVPR)},
  month     = {June},
  year      = {2025},
  pages     = {19424-19434}
}

@article{Qwen2.5-VL,
  title={Qwen2.5-VL Technical Report},
  author={Bai, Shuai and Chen, Keqin and Liu, Xuejing and Wang, Jialin and Ge, Wenbin and Song, Sibo and Dang, Kai and Wang, Peng and Wang, Shijie and Tang, Jun and Zhong, Humen and Zhu, Yuanzhi and Yang, Mingkun and Li, Zhaohai and Wan, Jianqiang and Wang, Pengfei and Ding, Wei and Fu, Zheren and Xu, Yiheng and Ye, Jiabo and Zhang, Xi and Xie, Tianbao and Cheng, Zesen and Zhang, Hang and Yang, Zhibo and Xu, Haiyang and Lin, Junyang},
  journal={arXiv preprint arXiv:2502.13923},
  year={2025}
}

@misc{qwen3technicalreport,
      title={Qwen3 Technical Report}, 
      author={Qwen Team},
      year={2025},
      eprint={2505.09388},
      archivePrefix={arXiv},
      primaryClass={cs.CL},
      url={https://arxiv.org/abs/2505.09388}, 
}

@article{wang2025internvl3,
  title={Internvl3. 5: Advancing open-source multimodal models in versatility, reasoning, and efficiency},
  author={Wang, Weiyun and Gao, Zhangwei and Gu, Lixin and Pu, Hengjun and Cui, Long and Wei, Xingguang and Liu, Zhaoyang and Jing, Linglin and Ye, Shenglong and Shao, Jie and others},
  journal={arXiv preprint arXiv:2508.18265},
  year={2025}
}

@article{li2024llava,
  	title={LLaVA-OneVision: Easy Visual Task Transfer},
  	author={Li, Bo and Zhang, Yuanhan and Guo, Dong and Zhang, Renrui and Li, Feng and Zhang, Hao and Zhang, Kaichen and Li, Yanwei and Liu, Ziwei and Li, Chunyuan},
  	journal={arXiv preprint arXiv:2408.03326},
  	year={2024}
}

@article{comanici2025gemini,
  title={Gemini 2.5: Pushing the frontier with advanced reasoning, multimodality, long context, and next generation agentic capabilities},
  author={Comanici, Gheorghe and Bieber, Eric and Schaekermann, Mike and Pasupat, Ice and Sachdeva, Noveen and Dhillon, Inderjit and Blistein, Marcel and Ram, Ori and Zhang, Dan and Rosen, Evan and others},
  journal={arXiv preprint arXiv:2507.06261},
  year={2025}
}

@article{achiam2023gpt,
  title={Gpt-4 technical report},
  author={Achiam, Josh and Adler, Steven and Agarwal, Sandhini and Ahmad, Lama and Akkaya, Ilge and Aleman, Florencia Leoni and Almeida, Diogo and Altenschmidt, Janko and Altman, Sam and Anadkat, Shyamal and others},
  journal={arXiv preprint arXiv:2303.08774},
  year={2023}
}

@misc{huang20253dr1enhancingreasoning3d,
      title={3D-R1: Enhancing Reasoning in 3D VLMs for Unified Scene Understanding}, 
      author={Ting Huang and Zeyu Zhang and Hao Tang},
      year={2025},
      eprint={2507.23478},
      archivePrefix={arXiv},
      primaryClass={cs.CV},
      url={https://arxiv.org/abs/2507.23478}, 
}

@InProceedings{Zhu_2025_ICCV,
    author    = {Zhu, Chenming and Wang, Tai and Zhang, Wenwei and Pang, Jiangmiao and Liu, Xihui},
    title     = {LLaVA-3D: A Simple yet Effective Pathway to Empowering LMMs with 3D Capabilities},
    booktitle = {Proceedings of the IEEE/CVF International Conference on Computer Vision (ICCV)},
    month     = {October},
    year      = {2025},
    pages     = {4295-4305}
}

@inproceedings{azuma_2022_CVPR,
  title={ScanQA: 3D Question Answering for Spatial Scene Understanding},
  author={Azuma, Daichi and Miyanishi, Taiki and Kurita, Shuhei and Kawanabe, Motoaki},
  booktitle={Proceedings of the IEEE/CVF Conference on Computer Vision and Pattern Recognition (CVPR)},
  year={2022}
}

@inproceedings{ma2022sqa3d,
  title={SQA3D: Situated Question Answering in 3D Scenes},
  author={Ma, Xiaojian and Yong, Silong and Zheng, Zilong and Li, Qing and Liang, Yitao and Zhu, Song-Chun and Huang, Siyuan},
  booktitle={International Conference on Learning Representations},
  year={2023},
  url={https://openreview.net/forum?id=IDJx97BC38}
}

@inproceedings{majumdar2023openeqa,
  author={Arjun Majumdar and Anurag Ajay and Xiaohan Zhang and Pranav Putta and Sriram Yenamandra and Mikael Henaff and Sneha Silwal and Paul Mcvay and Oleksandr Maksymets and Sergio Arnaud and Karmesh Yadav and Qiyang Li and Ben Newman and Mohit Sharma and Vincent Berges and Shiqi Zhang and Pulkit Agrawal and Yonatan Bisk and Dhruv Batra and Mrinal Kalakrishnan and Franziska Meier and Chris Paxton and Sasha Sax and Aravind Rajeswaran},
  title={{OpenEQA: Embodied Question Answering in the Era of Foundation Models}},
  booktitle={{CVPR}},
  year={2024},
}

@inproceedings{dai2017scannet,
    title={ScanNet: Richly-annotated 3D Reconstructions of Indoor Scenes},
    author={Dai, Angela and Chang, Angel X. and Savva, Manolis and Halber, Maciej and Funkhouser, Thomas and Nie{\ss}ner, Matthias},
    booktitle = {Proc. Computer Vision and Pattern Recognition (CVPR), IEEE},
    year = {2017}
}

@inproceedings{xu2024pointllm,
  title={PointLLM: Empowering Large Language Models to Understand Point Clouds},
  author={Xu, Runsen and Wang, Xiaolong and Wang, Tai and Chen, Yilun and Pang, Jiangmiao and Lin, Dahua},
  booktitle={ECCV},
  year={2024}
}

@InProceedings{Chen_2024_CVPR,
    author    = {Chen, Sijin and Chen, Xin and Zhang, Chi and Li, Mingsheng and Yu, Gang and Fei, Hao and Zhu, Hongyuan and Fan, Jiayuan and Chen, Tao},
    title     = {LL3DA: Visual Interactive Instruction Tuning for Omni-3D Understanding Reasoning and Planning},
    booktitle = {Proceedings of the IEEE/CVF Conference on Computer Vision and Pattern Recognition (CVPR)},
    month     = {June},
    year      = {2024},
    pages     = {26428-26438}
}

@article{tang2024minigpt,
  title={MiniGPT-3D: Efficiently Aligning 3D Point Clouds with Large Language Models using 2D Priors},
  author={Tang, Yuan and Han, Xu and Li, Xianzhi and Yu, Qiao and Hao, Yixue and Hu, Long and Chen, Min},
  journal={arXiv preprint arXiv:2405.01413},
  year={2024}
}

@article{hong20233d,
  title={3d-llm: Injecting the 3d world into large language models},
  author={Hong, Yining and Zhen, Haoyu and Chen, Peihao and Zheng, Shuhong and Du, Yilun and Chen, Zhenfang and Gan, Chuang},
  journal={Advances in Neural Information Processing Systems},
  volume={36},
  pages={20482--20494},
  year={2023}
}

@inproceedings{zhang2024chatscene,
  title={ChatScene: Knowledge-Enabled Safety-Critical Scenario Generation for Autonomous Vehicles},
  author={Zhang, Jiawei and Xu, Chejian and Li, Bo},
  booktitle={Proceedings of the IEEE/CVF Conference on Computer Vision and Pattern Recognition},
  pages={15459--15469},
  year={2024}
}

@inproceedings{yang2025thinking,
  title={Thinking in space: How multimodal large language models see, remember, and recall spaces},
  author={Yang, Jihan and Yang, Shusheng and Gupta, Anjali W and Han, Rilyn and Fei-Fei, Li and Xie, Saining},
  booktitle={Proceedings of the Computer Vision and Pattern Recognition Conference},
  pages={10632--10643},
  year={2025}
}

@article{wu2025spatial,
  title={Spatial-mllm: Boosting mllm capabilities in visual-based spatial intelligence},
  author={Wu, Diankun and Liu, Fangfu and Hung, Yi-Hsin and Duan, Yueqi},
  journal={arXiv preprint arXiv:2505.23747},
  year={2025}
}

@inproceedings{xu2024vlmgrounder,
  title={VLM-Grounder: A VLM Agent for Zero-Shot 3D Visual Grounding},
  author={Xu, Runsen and Huang, Zhiwei and Wang, Tai and Chen, Yilun and Pang, Jiangmiao and Lin, Dahua},
  booktitle={CoRL},
  year={2024}
}

@article{chen2024grounded3dllm,
      title={Grounded 3D-LLM with Referent Tokens}, 
      author={Chen, Yilun and Yang, Shuai and Huang, Haifeng and Wang, Tai and Lyu, Ruiyuan and Xu, Runsen and Lin, Dahua and Pang, Jiangmiao},
      journal={arXiv preprint arXiv:2405.10370},
      year={2024},
}

@article{wang2023chat,
  title={Chat-3d: Data-efficiently tuning large language model for universal dialogue of 3d scenes},
  author={Wang, Zehan and Huang, Haifeng and Zhao, Yang and Zhang, Ziang and Zhao, Zhou},
  journal={arXiv preprint arXiv:2308.08769},
  year={2023}
}

@InProceedings{Hu_2025_CVPR,
    author    = {Hu, Kai and Gao, Feng and Nie, Xiaohan and Zhou, Peng and Tran, Son and Neiman, Tal and Wang, Lingyun and Shah, Mubarak and Hamid, Raffay and Yin, Bing and Chilimbi, Trishul},
    title     = {M-LLM Based Video Frame Selection for Efficient Video Understanding},
    booktitle = {Proceedings of the IEEE/CVF Conference on Computer Vision and Pattern Recognition (CVPR)},
    month     = {June},
    year      = {2025},
    pages     = {13702-13712}
}

@inproceedings{huang2024embodied,
  title={An Embodied Generalist Agent in 3D World},
  author={Huang, Jiangyong and Yong, Silong and Ma, Xiaojian and Linghu, Xiongkun and Li, Puhao and Wang, Yan and Li, Qing and Zhu, Song-Chun and Jia, Baoxiong and Huang, Siyuan},
  booktitle={Proceedings of the International Conference on Machine Learning (ICML)},
  year={2024}
}

@article{GPT4Scene,
  title={GPT4Scene: Understand 3D Scenes from Videos with Vision-Language Models},
  author={Zhangyang Qi and Zhixiong Zhang and Ye Fang and Jiaqi Wang and Hengshuang Zhao},
  journal={arXiv preprint arXiv:2501.01428},
  year={2024}
}

@inproceedings{lin2024video,
  title={Video-llava: Learning united visual representation by alignment before projection},
  author={Lin, Bin and Ye, Yang and Zhu, Bin and Cui, Jiaxi and Ning, Munan and Jin, Peng and Yuan, Li},
  booktitle={Proceedings of the 2024 Conference on Empirical Methods in Natural Language Processing},
  pages={5971--5984},
  year={2024}
}

@InProceedings{Liu_2025_CVPR,
    author    = {Liu, Benlin and Dong, Yuhao and Wang, Yiqin and Ma, Zixian and Tang, Yansong and Tang, Luming and Rao, Yongming and Ma, Wei-Chiu and Krishna, Ranjay},
    title     = {Coarse Correspondences Boost Spatial-Temporal Reasoning in Multimodal Language Model},
    booktitle = {Proceedings of the IEEE/CVF Conference on Computer Vision and Pattern Recognition (CVPR)},
    month     = {June},
    year      = {2025},
    pages     = {3783-3792}
}

@inproceedings{chen2020scanrefer,
    title={Scanrefer: 3d object localization in rgb-d scans using natural language},
    author={Chen, Dave Zhenyu and Chang, Angel X and Nie{\ss}ner, Matthias},
    booktitle={Computer Vision--ECCV 2020: 16th European Conference, Glasgow, UK, August 23--28, 2020, Proceedings, Part XX 16},
    pages={202--221},
    year={2020},
    organization={Springer}
}

@InProceedings{Zhang_2023_ICCV,
    author    = {Zhang, Yiming and Gong, ZeMing and Chang, Angel X.},
    title     = {Multi3DRefer: Grounding Text Description to Multiple 3D Objects},
    booktitle = {Proceedings of the IEEE/CVF International Conference on Computer Vision (ICCV)},
    month     = {October},
    year      = {2023},
    pages     = {15225-15236}
}

@inproceedings{
dehghan2021arkitscenes,
title={{ARK}itScenes - A Diverse Real-World Dataset for 3D Indoor Scene Understanding Using Mobile {RGB}-D Data},
author={Gilad Baruch and Zhuoyuan Chen and Afshin Dehghan and Tal Dimry and Yuri Feigin and Peter Fu and Thomas Gebauer and Brandon Joffe and Daniel Kurz and Arik Schwartz and Elad Shulman},
booktitle={Thirty-fifth Conference on Neural Information Processing Systems Datasets and Benchmarks Track (Round 1)},
year={2021},
url={https://openreview.net/forum?id=tjZjv_qh_CE}
}

@inproceedings{yeshwanth2023scannet++,
  title={Scannet++: A high-fidelity dataset of 3d indoor scenes},
  author={Yeshwanth, Chandan and Liu, Yueh-Cheng and Nie{\ss}ner, Matthias and Dai, Angela},
  booktitle={Proceedings of the IEEE/CVF International Conference on Computer Vision},
  pages={12--22},
  year={2023}
}
\bibliographystyle{iclr2026_conference}
\appendix
\section{Appendix}
\label{sec:supp_outline}

In this supplementary material, we provide more details in the following sections:
\begin{itemize}
    \item 
    In~\cref{sec:supp_categorical_openeqa}, we present the categorical-level performance comparison between our method and the uniform sampling baseline, using the OpenEQA-HM3D dataset~\citep{majumdar2023openeqa}. 
    The results in this experiment further corroborates the benefits of our solution by archiving superior performance while using comparable amount of tokens.
    \item In~\cref{sec:supp_vsibench}, we report categorical-level results on VSI-Bench~\citep{yang2025thinking}, a benchmark designed to evaluate visual-spatial intelligence, including spatial size estimation and spatial relationship reasoning. The experimental results demonstrate that our method generalizes beyond 3D question answering benchmarks such as ScanQA and SQA3D, and remains effective on visual-spatial intelligence benchmarks.
    \item In~\cref{sec:supp_pruning_threshold}, we provide more details on the pruning threshold $\tau_o$ 
    by comparing results with differentnumbers of frames as the baseline and report categorical-level performance on the SQA3D dataset~\citep{ma2022sqa3d}. 
    \item In~\cref{sec:supp_scaling_frames}, we present more details on scaling the number of frames by reporting categorical-level performance across different model sizes and compared to uniform sampling on SQA3D. 
    \item In~\cref{sec:additional_qualitative_results}, we provide additional qualitative results on OpenEQA-HM3D and SQA3D.
\end{itemize}

\section{Categorical results on OpenEQA-HM3D.}
\label{sec:supp_categorical_openeqa}
\begin{table*}[ht]
\centering
\fontsize{6}{11}\selectfont
\centering
\caption{\textbf{Categorical results on OpenEQA-HM3D.} Comparisons on online sampling strategies, with evaluation results reported across seven categories of OpenEQA-HM3D.}
\vspace{1mm}
\begin{tabular}{l c cccccccc c}
    \toprule
    \multirow{4}{*}{Model} 
    & \multirow{2}{*}{Sampling}
    & \multicolumn{8}{c}{OpenEQA-HM3D~\cite{majumdar2023openeqa}}
    \\
    \cmidrule(lr){3-10}& \multirow{2}{*}{Strategy} 
    &Attr.&Func.&Obj.&Obj.&Obj. State&Spatial&World&\multirow{2}{*}{LLM Match$~\uparrow$}& \multirow{2}{*}{Tokens$~\downarrow$}\\
    &&
    Recog.&Reason.&Local.&Recog.&Recog.&Under.&Know.&&\\
    \midrule

    \multirow{2}{*}{Qwen2.5-VL-7B} 
& Uniform
    & 63.2 & 61.4 & 51.9 & 49.3 & 48.6 & 45.1 & \textbf{53.1} & 53.1 & \textbf{4.4M} \\
& Uniform + Ours
    & \textbf{71.6} & \textbf{61.8} & \textbf{54.4} & \textbf{54.5} & \textbf{63.6} & \textbf{49.6} & {51.2} & \textbf{58.2} & 5.2M \\
    
    \midrule
    
    \multirow{2}{*}{Qwen3-VL-8B} 
& Uniform
    & \textbf{71.4} & 66.4 & 70.2 & 64.0 & 66.5 & \textbf{59.8} & 70.0 & 67.1 & \textbf{3.4M}\\
& Uniform + Ours
    & 69.6 & \textbf{72.5} & \textbf{70.2} & \textbf{65.5} & \textbf{70.4} & 53.8 & \textbf{70.7} & \textbf{67.8} & {4.1M} \\
    
    \bottomrule
\end{tabular}

\label{tab_hm3d_category}
\vspace{3mm}
\end{table*}

In~\cref{tab_hm3d_category}, we report categorical-level results on the OpenEQA-HM3D dataset. The questions are categorized into seven types: (1) Attribute Recognition, (2) Functional Reasoning, (3) Object Localization, (4) Object Recognition, (5) Object State Recognition, (6) Spatial Understanding, and (7) World Knowledge. With comparable token usage, our method on top of uniform sampling yields better overall LLM Match scores. Our method demonstrates consistent improvements across nearly all categories for both Qwen3-VL and Qwen2.5-VL models. Notably, our method achieves a 15-point improvement in LLM Match score for the ``Object State Recognition'' category with Qwen2.5-VL-7B, while also showing meaningful gains in other categories such as Attribute Recognition with an 8.4-point improvement.


\section{Categorical results on VSI-Bench.}
\label{sec:supp_vsibench}
\begin{table}[ht]
\centering
\fontsize{6.5}{11}\selectfont
\centering
\caption{\textbf{Categorical results on VSI-Bench.} Comparisons on online sampling strategies, with evaluation results reported across seven categories of VSI-Bench.}
\vspace{1mm}
\begin{tabular}{l c ccccccccc c}
    \toprule
    \multirow{4}{*}{Model} 
    & \multirow{2}{*}{Sampling}
    & \multicolumn{9}{c}{VSIBench (ScanNet)}
    \\
    \cmidrule(lr){3-10}& \multirow{2}{*}{Strategy} 
    &Obj.&Abs.&Obj.&Room&Rel.&Rel.&Route&Appr.&\multirow{2}{*}{Avg.$~\uparrow$}& \multirow{2}{*}{Tokens$~\downarrow$}\\
    &&
    Count&Dist.&Size&Size&Dist.&Dir.&Plan&Order&\\
    \midrule

    \multirow{2}{*}{Qwen2.5-VL-7B} 
& Uniform
    & 33.7 & 9.1 & \textbf{31.4} & 36.7 & 40.5 & 36.4 & \textbf{25.0} & 28.2 & 29.3 & 16.4M \\
& Uniform + Ours
    & \textbf{34.3} & \textbf{9.5} & 31.1 & \textbf{40.5} & \textbf{41.3} & \textbf{39.4} & 23.4 & \textbf{30.3} &\textbf{30.5} & \textbf{14.8M} \\
    
    \midrule
    
    \multirow{2}{*}{Qwen3-VL-8B} 
& Uniform
    & 66.3 & 47.1 & 69.4 & 53.2 & 54.1 & 46.6 & 32.8 & 52.1 & 54.5 & 12.7M\\
& Uniform + Ours
    & \textbf{73.4} & \textbf{44.8} & \textbf{70.0} & \textbf{54.8} & \textbf{57.1} & \textbf{51.9} & 32.8 &\textbf{61.3} & \textbf{58.1} & \textbf{11.9M} \\
    
    \bottomrule
\end{tabular}

\label{tab_vsi}
\vspace{3mm}
\end{table}

In~\cref{tab_vsi}, we report categorical-level results on the VSI-Bench. The benchmark consists of three datasets: ScanNet~\citep{dai2017scannet}, ScanNet++~\citep{yeshwanth2023scannet++}, and ARKitScenes~\citep{dehghan2021arkitscenes}. We evaluate on the ScanNet split, where questions are categorized into eight types: (1) Object Count, (2) Absolute Distance, (3) Object Size, (4) Room Size, (5) Relative Distance, (6) Relative Direction, (7) Route Plan, and (8) Appearance Order. Following the original VSI-Bench metric, the first four categories require numerical answers while the latter four are multi-choice questions. Our method applied on top of uniform sampling achieves better average scores for both models. Notably, for Qwen3-VL-8B, our approach improves the average score from 54.5 to 58.1 while reducing token usage from 12.7M to 11.9M.

\section{More Details on the Pruning Threshold.}
\label{sec:supp_pruning_threshold}
\begin{table}[h]
\centering
\fontsize{6}{11}\selectfont

\centering
\caption{\textbf{More details on the pruning threshold~$\tau_o$.} We report six categorical results on SQA3D to investigate how different threshold settings affect pruning behavior, with higher $\tau_o$ values leading to less aggressive token removal. The results show that a threshold of 100\% performs the best across different baselines. ``F'' denotes the number of frames sampled.}
\vspace{1mm}
\begin{tabular}{l c c ccccccc c}
    \toprule
    \multirow{2}{*}{Model} 
    & \multirow{1}{*}{Sampling}
    & \multirow{2}{*}{$\tau_o$}
    & \multicolumn{8}{c}{SQA3D~\cite{ma2022sqa3d}}
    \\
    \cmidrule(lr){4-11}
    &Strategy&&what&which&can&is&how&others&Average $~\uparrow$& Tokens$~\downarrow$\\
    \midrule
    \multirow{10}{*}{Qwen3-VL-8B} 
    
    & Uniform 70F & - & 46.6 & 45.9 & 54.7 & 62.9 & 47.4 & 51.9 & 51.4 & 74.5M \\
    \cmidrule{2-11}
    & Uniform 70F + Ours& 25\% 
    & 45.9 & 43.6 & 52.7 & 61.2 & 43.9 & 50.5 & 49.6 & \textbf{10.7M} \\
    & Uniform 70F + Ours & 50\% 
    & 46.8 & 43.9 & 52.7 & 63.5 & 45.8 & 51.8 & 50.8 & 13.5M\\
    & Uniform 70F + Ours & 80\% 
    & 46.6 & 44.7 & \textbf{57.7} & 61.0 & 45.2 & 51.6 & 50.8 & 19.4M \\
    & Uniform 70F + Ours & 100\% 
    & \textbf{48.3} & \textbf{47.0} & 56.2 & \textbf{70.0} & \textbf{49.0} & \textbf{52.1} & \bf{52.2} & 35.2M\\ 
    \cmidrule[0.8pt]{2-11}
    & Uniform 20F & - & 45.6 & 45.6 & 54.1 & 60.6 & 46.2 & \textbf{51.1} & 50.2 & 21.6M \\
    \cmidrule{2-11}
    & Uniform 20F + Ours& 25\% 
    & 44.6 & \textbf{46.4} & 55.9 & 60.4 & 43.2 & 48.9 & 49.3 & \textbf{8.3M} \\
    & Uniform 20F + Ours & 50\% 
    & 44.6 & 43.0 & 55.3 & 61.2 & 43.9& 48.9 & 49.1 & 9.2M\\
    & Uniform 20F + Ours & 80\% 
    & 45.4 & 43.6 & 55.0 & 61.2 & 46.7 & 49.3 & 49.9 & 11.0M \\
    & Uniform 20F + Ours & 100\% 
    & \textbf{46.2} & 43.6 & \textbf{58.3} & \textbf{61.5} & \textbf{46.9} & 48.8 & \bf{50.5} & 14.6M\\ 
    \bottomrule
\end{tabular}


\label{tab_supp_pruning_threshold}
\end{table}

In~\cref{tab_supp_pruning_threshold}, we further investigate the effect of the pruning threshold $\tau_o$. 
From the results, we observe that applying our method on top of uniform sampling frames can significantly reduce token usage across different threshold settings. When $\tau_o$ is set to 25\%, the method aggressively reduces redundant information but incurs a performance drop of 1.8 points on the 70-frame baseline. This trend is consistent across both 70-frame and 20-frame baselines. \textbf{However, when our geometry-aware token pruning employs a milder strategy with $\tau_o$ set to 100\%, it removes only highly redundant information, achieving both token reduction and performance improvement}. Specifically, with $\tau_o$ set to 100\%, we achieve over 50\% token reduction on the 70-frame baseline and more than 30\% on the 20-frame baseline. The larger token reduction on the 70-frame baseline is due to denser frame sampling, which leads to more overlapping regions that can be pruned. This trend holds consistently across different uniform sampling frame baselines. Notably, applying our method with $\tau_o$ set to 100\% on the 70-frame baseline, the ``is'' category achieves a more than 7-point improvement in Exact Match score.

\section{More Details on Scaling Sampled Frames.}
\label{sec:supp_scaling_frames}
\begin{table*}[ht]
\centering
\fontsize{8}{14}\selectfont

\centering
\caption{\textbf{More details on scaling the number of frames.} We report six categorical results on SQA3D using different frame counts as baselines and apply our method with different model sizes. By adopting our method, nearly all baselines achieve performance gains while using fewer tokens. ``F'' denotes the number of frames sampled.}
\vspace{1mm}
\begin{tabular}{l c cccccccc}
    \toprule
    \multirow{2}{*}{Model} 
    & \multirow{1}{*}{Sampling}
    & \multicolumn{8}{c}{SQA3D~\cite{ma2022sqa3d}}
    \\
    \cmidrule(lr){3-10}
    &Strategy&what&which&can&is&how&others&Average $~\uparrow$& Tokens$~\downarrow$\\
    \midrule
    \multirow{8}{*}{Qwen3-VL-8B} 
    
    & Uniform 70F
    & 46.6 & 45.9 & 54.7 & \textbf{62.9} & 47.4 & 51.9 & 51.4 & 74.5M \\
    & + Ours
    & \textbf{48.3} & \textbf{47.0} & \textbf{56.2} & 62.0 & \textbf{49.0} & \textbf{52.1} & \textbf{52.2} \textcolor{mygreen}{\textbf{(+0.8)}}& \textbf{35.2M} \textcolor{mygreen}{\textbf{(-53\%)}}\\
    \cmidrule{2-10}
    & Uniform 50F 
    & 47.0 & \textbf{47.9} & \textbf{58.6} & \textbf{62.4} & \textbf{47.3} & 50.2 & \textbf{51.6} & 53.5M \\
    & + Ours
    & \textbf{47.7} & 47.0 & \textbf{58.6} & 61.4 & 46.2 & \textbf{51.2} & \textbf{51.6} & \textbf{27.6M} \textcolor{mygreen}{\textbf{(-45\%)}}\\
    \cmidrule{2-10}
    & Uniform 30F
    & 46.3 & 47.6 & 55.0 & \textbf{62.3} & \textbf{45.8} & 50.0 & \textbf{50.8} & 32.2M\\
    & + Ours
    & \textbf{46.4} & \textbf{48.2} & \textbf{55.6} & 60.9 & 44.3 & \textbf{51.4} & 50.7 \textcolor{red}{(-0.1)} & \textbf{19.4M} \textcolor{mygreen}{\textbf{(-40\%)}}\\
    \cmidrule{2-10}
    & Uniform 20F 
    & 45.6 & \textbf{45.6} & 54.1 & 60.6 & 46.2 & \textbf{51.1} & 50.2 & 21.6M \\
    & + Ours
    & \textbf{46.2} & 43.6 & \textbf{58.3} & \textbf{61.5} & \textbf{46.9} & 48.8 & \textbf{50.4} \textcolor{mygreen}{\textbf{(+0.2)}} & \textbf{14.6M} \textcolor{mygreen}{\textbf{(-32\%)}}\\

    \bottomrule
    \multirow{8}{*}{Qwen3-VL-4B} 
    
    & Uniform 70F
    & 45.5 & \textbf{46.4} & 60.4 & 60.7 & \textbf{52.3} & \textbf{48.8} & 51.3 & 74.5M \\
    & + Ours
    & \textbf{45.8} & 45.3 & \textbf{62.4} & \textbf{62.0} & 51.2 & 47.7 & \textbf{51.4} \textcolor{mygreen}{\textbf{(+0.1)}} & \textbf{35.2M} \textcolor{mygreen}{\textbf{(-53\%)}}\\
    \cmidrule{2-10}
    & Uniform 50F 
    & 45.4 & \textbf{47.0} & 59.8 & \textbf{62.3} & 48.4 & \textbf{48.8} & 51.0 & 53.5M \\
    & + Ours
    & \textbf{46.6} & 46.4 & \textbf{60.4} & 61.7 & \textbf{48.8} & 47.9 & \textbf{51.2} \textcolor{mygreen}{\textbf{(+0.2)}} & \textbf{27.6M} \textcolor{mygreen}{\textbf{(-45\%)}}\\
    \cmidrule{2-10}
    & Uniform 30F
    & 44.0 & 47.3 & 62.4 & 60.9 & 48.6 & \textbf{50.9} & 51.0 & 32.2M\\
    & + Ours
    & \textbf{45.0} & \textbf{48.4} & \textbf{63.6} & \textbf{62.3} & \textbf{49.9} & 50.0 & \textbf{51.8} \textcolor{mygreen}{\textbf{(+0.8)}} & \textbf{19.4M} \textcolor{mygreen}{\textbf{(-40\%)}}\\
    \cmidrule{2-10}
    & Uniform 20F 
    & 42.8 & \textbf{47.3} & \textbf{61.5} & 61.2 & 46.5 & 48.8 & 49.9 & 21.6M \\
    & + Ours
    & \textbf{43.9} & 46.2 & 61.0 & \textbf{61.5} & \textbf{46.9} & \textbf{50.0} & \textbf{50.4} \textcolor{mygreen}{\textbf{(+0.5)}} & \textbf{14.6M} \textcolor{mygreen}{\textbf{(-32\%)}}\\    
    
    \bottomrule
\end{tabular}


\label{tab_supp_scaling_frames}
\vspace{-0.3cm}
\end{table*}

Applying our geometry-aware pruning method on top of different uniform sampling strategies with models of different sizes, Qwen3-VL-8B and Qwen3-VL-4B, we report categorical-level performance on SQA3D in~\cref{tab_supp_scaling_frames}. We adopt increasing uniform sampling frame counts of 20, 30, 50, and 70 as baselines and apply our method on top of each. We observe that our method consistently achieves larger performance improvements than the baselines while using fewer tokens. For Qwen3-VL-8B, the baseline achieves an Exact Match of 51.4 with 70 uniformly sampled frames. Applying our method on the same 70 frames improves the Exact Match to 52.2 while consuming less than 50\% of the tokens. The same trend appears in the smaller model. From the categorical-level results, we observe that nearly all categories benefit from our method while using significantly fewer computational resources.
\section{Additional Qualitative Results.}
\label{sec:additional_qualitative_results}

\begin{figure*}[t]
    \centering
    \includegraphics[width=1.0\linewidth]{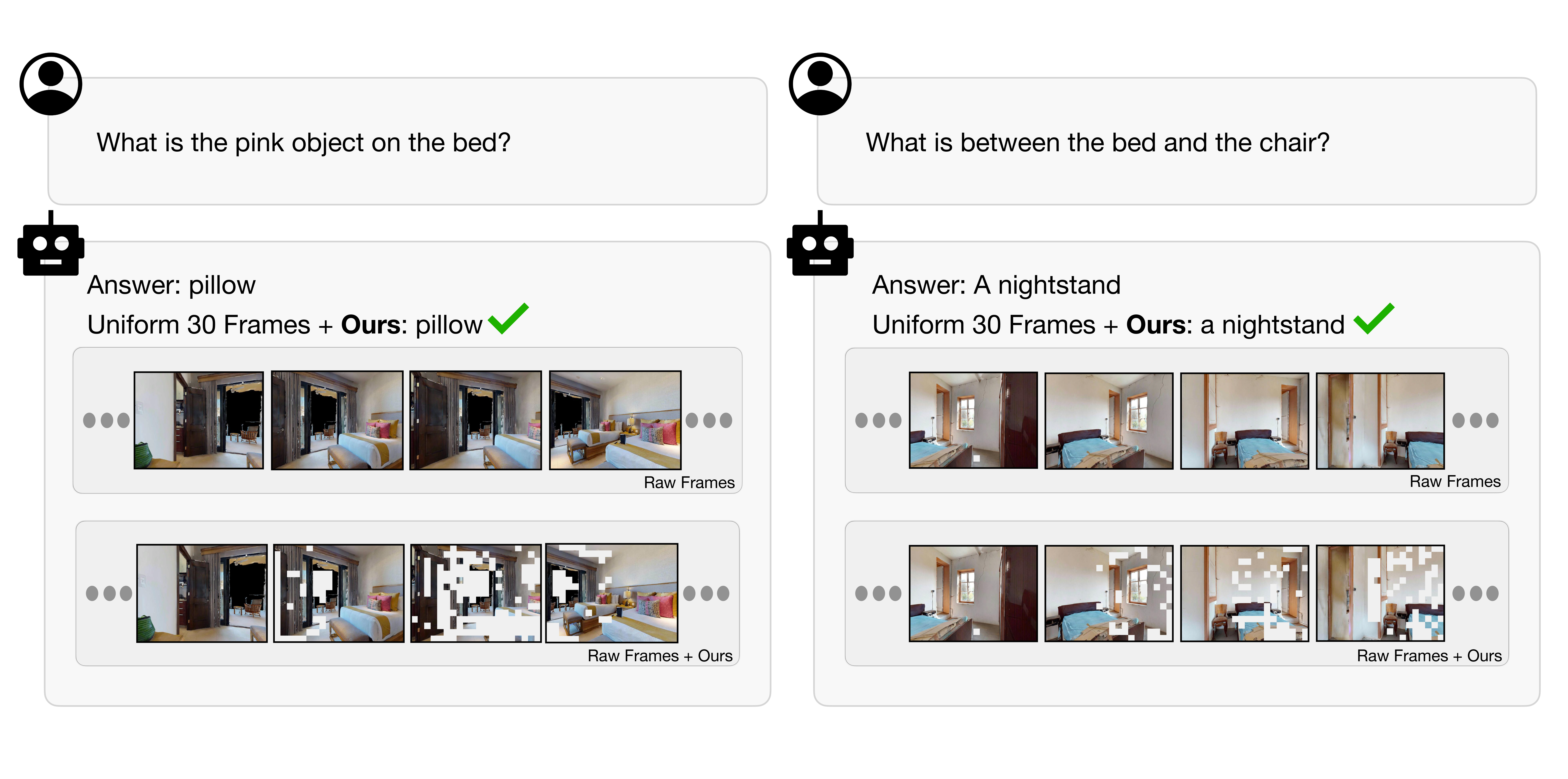}
    \caption{\textbf{Qualitative results on OpenEQA-HM3D.} We adopt our method with a uniform sampling strategy on OpenEQA-HM3D. The left and right examples show different scenes with different questions. Our method can effectively prune the overlapping regions.}
    
    \label{fig:supp_hm3d_qualitative_result}
\end{figure*}

\begin{figure*}[t]
    \centering
    \includegraphics[width=1\linewidth]{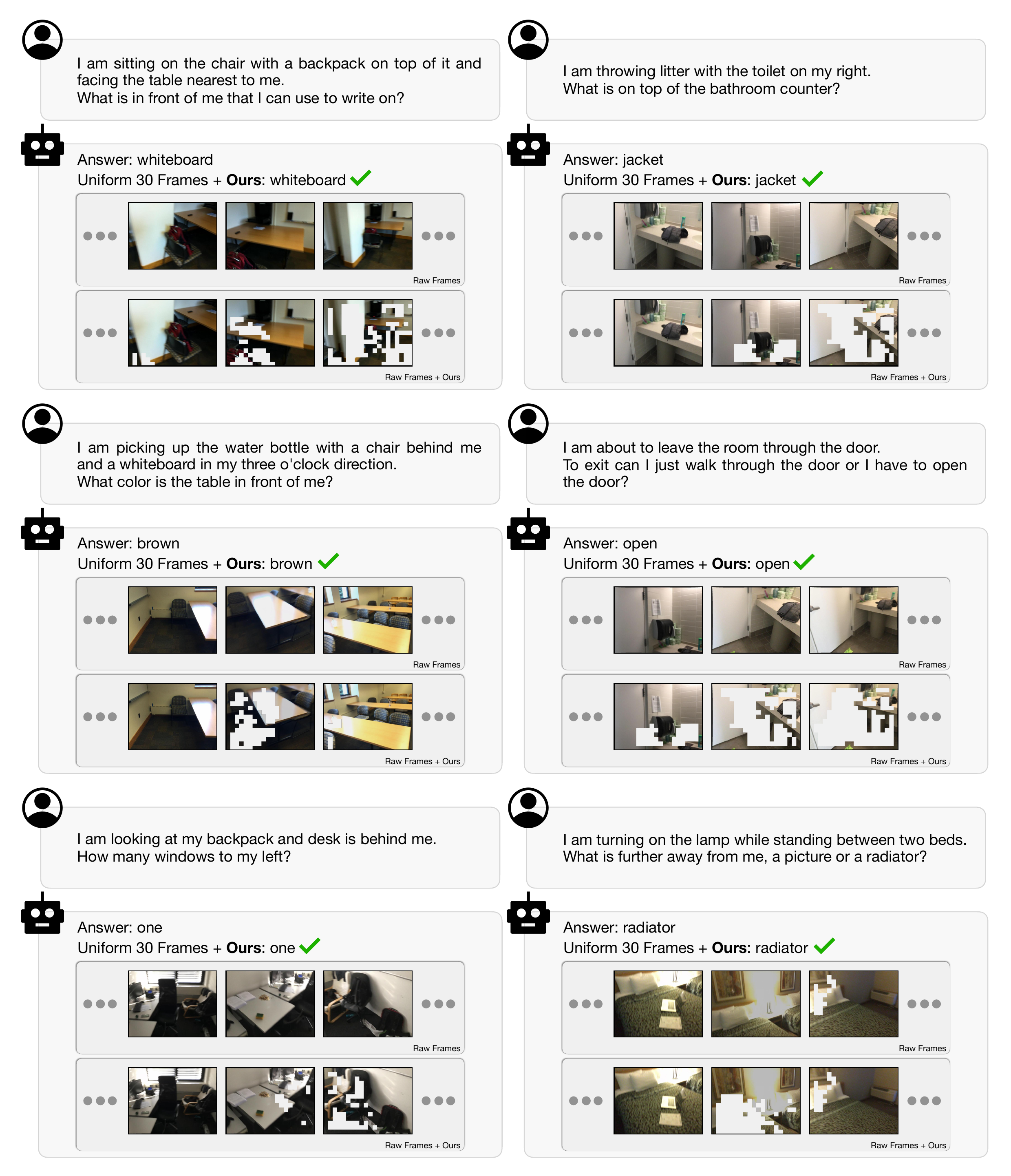}
    \caption{\textbf{Additional qualitative results.} We apply our method on top of uniformly sampled 30 frames on the SQA3D dataset. The qualitative results show that our method prunes redundant information while preserving important details, thereby improving both token efficiency and performance.}
    
    \label{fig:supp_more_qualitative_result}
\end{figure*}

We present additional qualitative results in~\cref{fig:supp_hm3d_qualitative_result} and~\cref{fig:supp_more_qualitative_result}. The top row shows the raw frames, and the bottom row visualizes the results after applying our pruning method. We use Qwen3-VL-8B as the base model for these results.

In~\cref{fig:supp_hm3d_qualitative_result}, we visualize two examples from the OpenEQA-HM3D dataset to illustrate how our method prunes redundant tokens and demonstrates its effectiveness. Our method removes highly overlapping information in an online manner, progressively pruning more tokens as additional frames are received because information may be repeated across frames. In the left example, we observe that in the third image, more regions are pruned as the same visual information has already appeared in previous frames. This pruning pattern is evident across both examples.

In~\cref{fig:supp_more_qualitative_result}, we provide six additional qualitative results from the SQA3D dataset, applying our method on top of uniformly sampled 30 frames. These examples illustrate a key limitation of uniform sampling: consecutive frames often contain highly redundant visual information. Our geometry-aware pruning method effectively removes this redundancy while preserving task-critical details. Importantly, our method maintains the model's ability to answer questions correctly while reducing token consumption and minimizing computational interference. By selectively pruning overlapping regions across frames, our method achieves a balance between computational efficiency and answer quality, which is particularly valuable in 3D question answering tasks.

\end{document}